# Towards Robust GNSS Positioning and Real-time Kinematic Using Factor Graph Optimization

Weisong Wen and Li-Ta Hsu*

*Abstract*— Global navigation satellite systems (GNSS) are one of the utterly popular sources for providing globally referenced positioning for autonomous systems. However, the performance of the GNSS positioning is significantly challenged in urban canyons, due to the signal reflection and blockage from buildings. Given the fact that the GNSS measurements are highly environmentally dependent and time-correlated, the conventional filtering-based method for GNSS positioning cannot simultaneously explore the time-correlation among historical measurements. As a result, the filtering-based estimator is sensitive to unexpected outlier measurements. In this paper, we present a factor graph-based formulation for GNSS positioning and real-time kinematic (RTK). The formulated factor graph framework effectively explores the time-correlation of pseudorange, carrier-phase, and doppler measurements, and leads to the non-minimal state estimation of the GNSS receiver. The feasibility of the proposed method is evaluated using datasets collected in challenging urban canyons of Hong Kong and significantly improved positioning accuracy is obtained, compared with the filtering-based estimator.

## I. INTRODUCTION

Global navigation satellite system (GNSS) [1] is currently one of the major sources for providing globally referenced positioning for autonomous systems with navigation requirements, such as the unmanned aerial vehicle (UAV) [2], autonomous driving vehicles (ADV) [3]. With the increased availability of multi-constellations, the GNSS solution becomes even more popular. In general, the major positioning methods involve GNSS positioning and GNSS real-time kinematic (RTK) positioning.

The popular GNSS positioning method is to use the extended Kalman filter (EKF) [4] to estimate the position, velocity, and time (PVT) of the GNSS receiver simultaneously based on the available GNSS measurements. General positioning accuracy (5~10 meters) [5] can be obtained in open sky areas. The remaining error is mainly caused by the ionosphere error, troposphere error and satellite clock/orbit biases, etc. To increase the accuracy of the GNSS positioning, RTK is proposed to perform GNSS positioning which can deliver centimeter-level positioning accuracy. The GNSS-RTK removes the errors (including the errors mentioned above and the receiver clock offset) using the double-difference technique based on the observations (e.g. pseudorange and carrier-phase measurements) received from a reference station. The GNSS-RTK positioning algorithm mainly includes two steps, the float solution estimation, and carrier-phase integer ambiguity resolution. A common approach [4] is to use an extended Kalman filter (EKF) [6] to estimate the float solution and the double-differenced (DD) carrier-phase ambiguity bias based on the DD pseudorange and carrier-phase measurements. Meanwhile, the LAMBDA algorithm [7] is employed to resolve the integer ambiguity to further achieve a fixed solution leading to centimeter-level accuracy. In short, the EKF dominates both the GNSS positioning and the GNSS-RTK positioning, due to the maturity and efficiency of the EKF estimator. Satisfactory performance can be obtained for GNSS-RTK (~5 centimeters) in open-sky areas where the error sources can be dealt with by differential techniques. Unfortunately, the performances of both the GNSS positioning and GNSS-RTK are significantly degraded in urban canyons [8] which are mainly due to the outlier measurements, arising from the multipath effects and none-line-of-sight (NLOS) [9] receptions caused by the building reflection and blockage. To mitigate the effects of GNSS outliers from NLOS receptions and multipath effects, numerous methods are studied, such as the 3D mapping aided GNSS (3DMA GNSS) [10-12], the 3D LiDAR aided GNSS positioning [13-16], and the camera aided GNSS positioning [1, 17]. However, these methods rely heavily on the availability of 3D mapping information or additional sensors.

Interestingly, instead of estimating the state of the GNSS receiver mainly based on the observation at the current epoch recursively via the EKF estimator, the recent researches [18-21] propose the factor graph-based formulation to process the GNSS pseudorange measurements and significantly improved performance is achieved, compared with the conventional EKF. The work [22] by a team from the Chemnitz University of Technology was the first paper utilizing factor graph optimization (FGO) in GNSS positioning. However, only the pseudorange measurements were considered. Then their continuous works focused on developing a robust model [23-25] for mitigating the effects of the potential NLOS receptions. Interestingly, a team from West Virginia University carried out similar researches [20, 26, 27], applying FGO models to GNSS precise point positioning (PPP) and obtaining significantly improved results. Inspired by the significant improvement arising from FGO, our previous work [28] extensively evaluated the performance of the integration of GNSS pseudorange and inertial measurement unit (IMU) using EKF and FGO. Our finding showed that the FGO could explore the time-correlation among the environment dependant GNSS pseudorange measurements simultaneously, leading to improved robustness against outliers, compared with the EKF-based estimator. However, the potential of FGO in

---
Weisong Wen and Li-Ta Hsu are with the Hong Kong Polytechnic University, Hong Kong. (corresponding author to provide e-mail: lt.hsu@polyu.edu.hk).

GNSS-RTK is not explored in the existing work. Meanwhile, the application of full suite GNSS measurements (pseudorange, Doppler and carrier-phase measurements) in FGO is not explored as well.

To fill this gap, this paper develops a factor graph-based formulation that provides the capability of GNSS positioning and GNSS-RTK positioning. Regarding the GNSS positioning, the pseudorange and Doppler measurements are integrated using FGO where the historical measurements are utilized simultaneously. Regarding the GNSS-RTK, the DD pseudorange and carrier-phase measurements are integrated using FGO where the states are connected using velocity estimated using Doppler measurements. To the best of the authors' knowledge, this is the first integrated framework that solves the GNSS positioning and GNSS-RTK using the state-of-the-art FGO. Meanwhile, we open-source the implementation of the proposed factor graph-based formulation [1]. The rest of the paper is summarized as follows: the derivation of FGO-based GNSS positioning and GNSS-RTK methods are described in Section III after the overview of the proposed method is presented in Section II. Then, the experiment is conducted to evaluate the performance of the proposed framework in Section IV. Finally, conclusions are drawn, together with future work.

## II. OVERVIEW OF THE PROPOSED METHOD

The architecture of the proposed pipeline is shown in Fig. 1 which mainly involves three segments: sensing, modeling, and optimization. The inputs of the system include the raw measurements from the GNSS receiver and the observations from the reference station for further GNSS-RTK positioning. The raw GNSS measurements mainly consist of the Doppler, pseudorange, and carrier-phase measurements. In the modeling segment, the velocity of the GNSS receiver is estimated based on the Doppler measurements using the least-squares (LS) algorithm. Meanwhile, the double-difference technique is employed to eliminate satellite and receiver clock bias, and atmospheric effects based on the observation from the reference station. In the optimization segment, the solution from GNSS positioning can be obtained by solving the factor graph formulated by velocity and pseudorange factor. Meanwhile, the float solution of GNSS-RTK can be estimated by solving the factor graph constructed by the DD pseudorange, DD carrier-phase, and Doppler measurements. Finally, the integer ambiguity is solved using the LAMBDA algorithm to achieve a fixed solution.

The contributions of this paper are listed as follows:

(1) This paper develops an FGO-based formulation, which provides the capability of GNSS positioning and RTK positioning. Meanwhile, additional sensor measurements, such as LiDAR odometry, can be easily integrated into our framework.

(2) We evaluate the performance of our framework using challenging datasets collected in the urban canyons of Hong Kong. The results show that the FGO-based GNSS positioning outperforms the one from the EKF-based solutions.

(3) The source code for the proposed framework is available online. Meanwhile, several datasets are also provided for algorithm development and evaluation of researchers. This is the first open-source implementation for GNSS positioning and GNSS-RTK using the FGO.

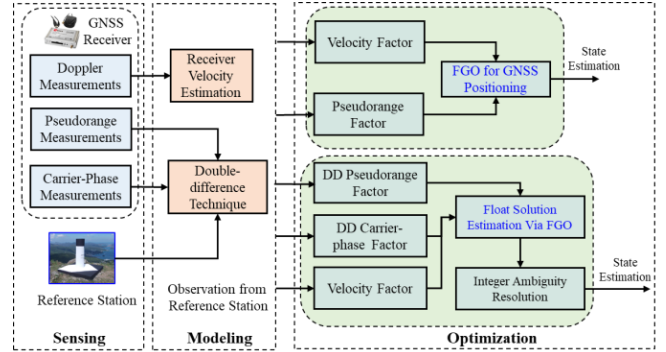

**Fig. 1** The architecture of the proposed pipeline. The "DD" stands for the double-difference technique.

To make the proposed pipeline clear, the following major notations are defined and followed by the rest of the paper. Be noted that the state of the GNSS receiver and the position of satellites are all expressed in the earth-centered, earth-fixed (ECEF) frame.

a) The pseudorange measurement received from a satellite $s$ at a given epoch $t$ is expressed as $\rho_{r,t}^s$. The subscript $r$ denotes the GNSS receiver. The superscript $s$ denotes the index of the satellite.
b) The Doppler measurement received from satellite $s$ at a given epoch $t$ is expressed as $d_{r,t}^s$.
c) The carrier-phase measurement received from a satellite $s$ at a given epoch $t$ is expressed as $\psi_{r,t}^s$.
d) The position of the satellite $s$ at a given epoch $t$ is expressed as $\mathbf{p}_t^s = (p_{t,x}^s, p_{t,y}^s, p_{t,z}^s)^T$.
e) The velocity of the satellite $s$ at a given epoch $t$ is expressed as $\mathbf{v}_t^s = (v_{t,x}^s, v_{t,y}^s, v_{t,z}^s)^T$.
f) The position of the GNSS receiver at a given epoch $t$ is expressed as $\mathbf{p}_{r,t} = (p_{r,t,x}, p_{r,t,y}, p_{r,t,z})^T$.
g) The velocity of the GNSS receiver at a given epoch $t$ is expressed as $\mathbf{v}_{r,t} = (v_{r,t,x}, v_{r,t,y}, v_{r,t,z})^T$.
h) The clock bias of the GNSS receiver at a given epoch $t$ is expressed as $\delta_{r,t}$, that with the unit in meters. $\delta_{r,t}^s$ denotes the satellite clock bias by meters.
i) The position of the base (reference) station is expressed as $\mathbf{p}_b = (p_{b,x}, p_{b,y}, p_{b,z})^T$. The variables $\rho_{b,t}^s$ and $\Phi_{b,t}^s$ denote the pseudorange and range measurements of carrier-phase from satellite $s$ received by the reference station at epoch $t$.

## III. METHODOLOGY

### A. GNSS Positioning Using Factor Graph Optimization

In terms of the measurements from the GNSS receiver, each pseudorange measurement, $\rho_{r,t}^s$, is denoted as follows [29].

---
[1] https://github.com/weisongwen/GraphGNSSLib

$$\rho_{r,t}^s = r_{r,t}^s + c(\delta_{r,t} - \delta_{r,t}^s) + I_{r,t}^s + T_{r,t}^s + \varepsilon_{r,t}^s \quad (1)$$

where $r_{r,t}^s$ is the geometric range between the satellite and the GNSS receiver. $I_{r,t}^s$ represents the ionospheric delay distance; $T_{r,t}^s$ indicates the tropospheric delay distance. $\varepsilon_{r,t}^s$ represents the errors caused by the multipath effects, NLOS receptions, receiver noise, antenna phase-related noise. In this paper, the satellite systems that we used include the global positioning system (GPS) and BeiDou. Besides, we follow the methods used in RTKLIB [4] to model the atmosphere effects ($I_{r,t}^s$ and $T_{r,t}^s$).

Given the Doppler measurement ($d_{r,t}^1, d_{r,t}^2, ...$) of each satellite at an epoch $t$, the velocity ($\mathbf{v}_{r,t}$) of the GNSS receiver can be calculated using the LS method [30]. Giving that the state of the velocity, $\mathbf{x}_t^d$, is as follows:

$$\mathbf{x}_t^d = (\mathbf{v}_{r,t}, \dot{\delta}_{r,t})^T \quad (2)$$

where the $\mathbf{v}_{r,t}$ represents the velocity of the GNSS receiver. The variable $\dot{\delta}_{r,t}$ stands for the receiver clock drift. The range rate measurement vector ($\mathbf{y}_{r,t}^d$) at an epoch $t$ is expressed as follows:

$$\mathbf{y}_{r,t}^d = (\lambda d_{r,t}^1, \lambda d_{r,t}^2, \lambda d_{r,t}^3, ...)^T \quad (3)$$

where the $\lambda$ denotes the carrier wavelength of the satellite signal, the $d_{r,t}^s$ represents the Doppler measurement. The observation function $h^d(*)$ which connects the state and the Doppler measurements are expressed as follows:

$$h^d(\mathbf{x}_t^d) = \begin{bmatrix} rr_{r,t}^1 + \dot{\delta}_{r,t} - \dot{\delta}_{r,t}^1 \\ rr_{r,t}^2 + \dot{\delta}_{r,t} - \dot{\delta}_{r,t}^2 \\ rr_{r,t}^3 + \dot{\delta}_{r,t} - \dot{\delta}_{r,t}^3 \\ ... \\ rr_{r,t}^m + \dot{\delta}_{r,t} - \dot{\delta}_{r,t}^m \end{bmatrix} \quad (4)$$

With $\mathbf{y}_{r,t}^d = h^d(\mathbf{x}_t^d) + \boldsymbol{\omega}_{r,t}^d$

where the superscript $m$ denotes the total number of satellites and the variable $\omega_{r,t}^d$ stands for the noise associated with the $\mathbf{y}_{r,t}^d$. The variable $rr_{r,t}^m$ denotes the expected range rate. The variables $\dot{\delta}_{r,t}$ and $\dot{\delta}_{r,t}^m$ represent the receiver and satellite clock bias drift. The Jacobian matrix $\mathbf{H}_t^d$ for the observation function $h^d(*)$ is denoted as follows:

$$\mathbf{H}_t^d = \begin{bmatrix} \frac{p_{t,x}^1 - p_{r,t,x}}{\|\mathbf{p}_t^1 - \mathbf{p}_{r,t}\|} & \frac{p_{t,y}^1 - p_{r,t,y}}{\|\mathbf{p}_t^1 - \mathbf{p}_{r,t}\|} & \frac{p_{t,z}^1 - p_{r,t,z}}{\|\mathbf{p}_t^1 - \mathbf{p}_{r,t}\|} & 1 \\ \frac{p_{t,x}^2 - p_{r,t,x}}{\|\mathbf{p}_t^2 - \mathbf{p}_{r,t}\|} & \frac{p_{t,y}^2 - p_{r,t,y}}{\|\mathbf{p}_t^2 - \mathbf{p}_{r,t}\|} & \frac{p_{t,z}^2 - p_{r,t,z}}{\|\mathbf{p}_t^2 - \mathbf{p}_{r,t}\|} & 1 \\ \frac{p_{t,x}^3 - p_{r,t,x}}{\|\mathbf{p}_t^3 - \mathbf{p}_{r,t}\|} & \frac{p_{t,y}^3 - p_{r,t,y}}{\|\mathbf{p}_t^3 - \mathbf{p}_{r,t}\|} & \frac{p_{t,z}^3 - p_{r,t,z}}{\|\mathbf{p}_t^3 - \mathbf{p}_{r,t}\|} & 1 \\ ... & ... & ... & ... \\ \frac{p_{t,x}^m - p_{r,t,x}}{\|\mathbf{p}_t^m - \mathbf{p}_{r,t}\|} & \frac{p_{t,y}^m - p_{r,t,y}}{\|\mathbf{p}_t^m - \mathbf{p}_{r,t}\|} & \frac{p_{t,z}^m - p_{r,t,z}}{\|\mathbf{p}_t^m - \mathbf{p}_{r,t}\|} & 1 \end{bmatrix} \quad (5)$$

where the operator $\|*\|$ is employed to calculate the range distance between the given satellite and the GNSS receiver. The expected range rate $rr_{r,t}^s$ for satellite $s$ can also be calculated as follows:

$$rr_{r,t}^s = \mathbf{e}_{r,t}^{s,LOS}(\mathbf{v}_t^s - \mathbf{v}_{r,t}) + \frac{\omega_{earth}}{c_L}(v_{t,y}^s p_{r,t,x} + p_{t,y}^s v_{r,t,x} - p_{t,x}^s v_{r,t,y} - v_{t,x}^s p_{r,t,y}) \quad (6)$$

where the variable $\omega_{earth}$ denotes the angular velocity of the earth rotation [4]. The variable $c_L$ denotes the speed of the light. The variable $\mathbf{e}_{r,t}^{s,LOS}$ denotes the line-of-sight vector connecting the GNSS receiver and the satellite (See equation (5)). Therefore, the velocity ($\mathbf{v}_{r,t}$) of the GNSS receiver can be estimated via LS [4] based on equations (4) and (5).

The graph structure of the proposed factor graph for solving the GNSS positioning is shown in Fig. 2. The subscript $n$ denotes the total epochs of measurements considered in the FGO. Each state in the factor graph is connected using the Doppler velocity factor. The state of the GNSS receiver is represented as follows:

$$\chi = [\mathbf{x}_{r,1}, \mathbf{x}_{r,2}, ..., \mathbf{x}_{r,n}] \quad (7)$$
$$\mathbf{x}_{r,t} = (\mathbf{p}_{r,t}, \mathbf{v}_{r,t}, \delta_{r,t})^T \quad (8)$$

where the variable $\chi$ denotes the set states of the GNSS receiver from the first epoch to the current $n$. The $\mathbf{x}_{r,t}$ denotes the state of the GNSS receiver at epoch $t$ which involves the position ($\mathbf{p}_{r,t}$), velocity ($\mathbf{v}_{r,t}$) and receiver clock bias ($\delta_{r,t}$).

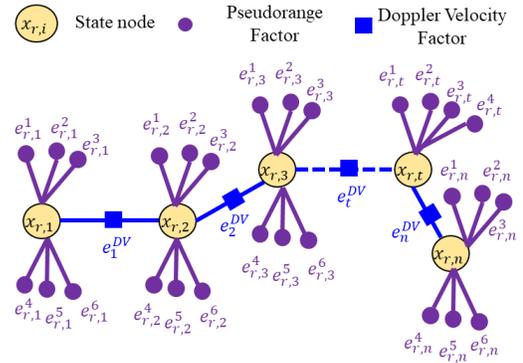

Fig. 2 The purple circle denotes the pseudorange factor (e.g $e_{r,t}^s$). The blue shaded rectangle represents the Doppler velocity factor (e.g $e_t^{DV}$). The yellow shaded circle stands for the state of the GNSS receiver.

The observation model for GNSS pseudorange measurement from a given satellite $s$ is represented as follows:

$$\rho_{r,t}^s = h_{r,t}^s(\mathbf{p}_{r,t}, \mathbf{p}_t^s, \delta_{r,t}) + \omega_{r,t}^s \quad (9)$$

with $h_{r,t}^s(\mathbf{p}_{r,t}, \mathbf{p}_t^s, \delta_{r,t}) = \|\mathbf{p}_t^s - \mathbf{p}_{r,t}\| + \delta_{r,t}$

where the variable $\omega_{r,t}^s$ stands for the noise associated with the $\rho_{r,t}^s$. Therefore, we can get the error function ($\mathbf{e}_{r,t}^s$) for a given satellite measurement $\rho_{r,t}^s$ as follows:

$$\|\mathbf{e}_{r,t}^s\|_{\Sigma_{r,t}^s}^2 = \|\rho_{r,t}^s - h_{r,t}^s(\mathbf{p}_{r,t}, \mathbf{p}_t^s, \delta_{r,t})\|_{\Sigma_{r,t}^s}^2 \quad (10)$$

where $\Sigma_{r,t}^s$ denotes the covariance matrix. We calculate the $\Sigma_{r,t}^s$ based on the satellite elevation angle, signal, and noise ratio (SNR) following the work in [31]. The observation model for the velocity ($\mathbf{v}_{r,t}$) is expressed as follows:

$$\mathbf{v}_{r,t}^{DV} = h_{r,t}^{DV}(\mathbf{x}_{r,t+1}, \mathbf{x}_{r,t}) + \boldsymbol{\omega}_{r,t}^{DV} \quad (11)$$

$$\text{with } h_{r,t}^{DV}(\mathbf{x}_{r,t+1}, \mathbf{x}_{r,t}) = \begin{bmatrix} (p_{r,t+1,x} - p_{r,t,x})/\Delta t \\ (p_{r,t+1,y} - p_{r,t,y})/\Delta t \\ (p_{r,t+1,z} - p_{r,t,z})/\Delta t \end{bmatrix}$$

where the $\mathbf{z}_{r,t}^{DV}$ denotes the velocity measurements given by the estimation in (2). The variable $\boldsymbol{\omega}_{r,t}^{DV}$ denotes the noise associated with the velocity measurement. The variable $\Delta t$ denotes the time difference between epoch $t$ and epoch $t+1$. Therefore, we can get the error function ($\mathbf{e}_{r,t}^{DV}$) for a given velocity measurement $\mathbf{v}_{r,t}^{DV}$ as follows:

$$||\mathbf{e}_{r,t}^{DV}||_{\Sigma_{r,t}^{DV}}^2 = ||\mathbf{v}_{r,t}^{DV} - h_{r,t}^{DV}(\mathbf{x}_{r,t+1}, \mathbf{x}_{r,t})||_{\Sigma_{r,t}^{DV}}^2 \qquad (12)$$

where $\Sigma_{r,t}^{DV}$ denotes the covariance matrix. Therefore, the objective function for the GNSS positioning using FGO is formulated as follows:

$$\boldsymbol{\chi}^* = \arg\min_{\boldsymbol{\chi}} \sum_{s,t} ||\mathbf{e}_{r,t}^{DV}||_{\Sigma_{r,t}^{DV}}^2 + ||\mathbf{e}_{r,t}^s||_{\Sigma_{r,t}^s}^2 \qquad (13)$$

The variable $\boldsymbol{\chi}^*$ denotes the optimal estimation of the state sets, which can be estimated by solving the objective function (13).

*B. GNSS-RTK Using Factor Graph Optimization*

In terms of the measurements from the GNSS receiver, each carrier-phase measurement, $\psi_{r,t}^s$, is written as follows [29].

$$\lambda \psi_{r,t}^s = r_{r,t}^s + c_L(\delta_{r,t} - \delta_{r,t}^s) + I_{r,t}^s + T_{r,t}^s + \lambda B_{r,t}^s + d\psi_{r,t}^s + \epsilon_{r,t}^s \qquad (14)$$

$$\text{where } B_{r,t}^s = \psi_{r,0,t} - \psi_{0,t}^s + N_{r,t}^s$$

where $B_{r,t}^s$ is the carrier-phase bias. The variable $\lambda$ denotes the carrier wavelength. The variable $d\psi_{r,t}^s$ denotes the carrier-phase correction term including antenna phase offsets and variations, station displacement by earth tides, phase windup effect, and relativity correction on the satellite clock. The detailed formulation of the carrier-phase correction can be found in [4]. The variable $\psi_{r,0,t}$ represents the initial phase of the receiver local oscillator. Similarly, the $\psi_{0,t}^s$ stands for the initial phase of the transmitted navigation signal from the satellite. The variable $N_{r,t}^s$ denotes the carrier-phase integer ambiguity. $\epsilon_{r,t}^s$ represents the errors caused by the multipath effects, NLOS receptions, receiver noise, antenna delay.

The DD pseudorange measurement ($\rho_{DD,t}^s$) of satellite $s$ is formulated as follows [4]:

$$\rho_{DD,t}^s = (\rho_{r,t}^s - \rho_{b,t}^s) - (\rho_{r,t}^w - \rho_{b,t}^w) \qquad (15)$$

The variable $\rho_{b,t}^w$ and $\rho_{b,t}^s$ stands for the pseudorange measurements received by the reference station which is denoted by the subscript "$b$". Generally, the satellite with the highest elevation angle tends to involve the lowest multipath and NLOS errors. Therefore, the satellite $w$, with the highest elevation angle, is selected as the master satellite. After applying the DD process to the pseudorange measurements, the derived $\rho_{DD,t}^s$ is free of the clock bias and atmosphere effects [4]. Similarly, the DD carrier-phase measurement ($\Phi_{DD,t}^s$) of satellite $s$ is formulated as follows [4]:

$$\Phi_{DD,t}^s = (\Phi_{r,t}^s - \Phi_{b,t}^s) - (\Phi_{r,t}^w - \Phi_{b,t}^w) \qquad (16)$$

The variables $\Phi_{b,t}^s$ and $\Phi_{b,t}^w$ stand for the carrier-phase measurements received by the reference station. Similarly, the clock bias and atmosphere effects are waived from $\Phi_{DD,t}^s$. Meanwhile, the $\Phi_{DD,t}^s$ involves the DD ambiguity [4], which is to be estimated.

The state of the GNSS receiver is represented as follows:

$$\boldsymbol{\chi} = [\mathbf{x}_{r,1}, \mathbf{x}_{r,2}, \dots, \mathbf{x}_{r,n}] \qquad (17)$$

$$\mathbf{x}_{r,t} = (\mathbf{p}_{r,t}, \mathbf{v}_{r,t}, \delta_{r,t}, \Delta N_{rb,t}^1, \Delta N_{rb,t}^2, \dots, \Delta N_{rb,t}^{m-1})^T \qquad (18)$$

where the variable $\boldsymbol{\chi}$ denotes the set states of the GNSS receiver from the first epoch to the current $n$. The $\mathbf{x}_{r,t}$ denotes the state of the GNSS receiver at epoch $t$ which consists of position ($\mathbf{p}_{r,t}$), velocity ($\mathbf{v}_{r,t}$) and the DD ambiguities. The variable $\Delta N_{rb,t}^{m-1}$ denotes the DD carrier-phase ambiguity bias corresponding to satellite $m-1$. In other words, each DD carrier-phase measurement involves a specific ambiguity bias. The graph structure for estimating the float solution of the GNSS-RTK is shown in Fig. 3. The yellow shaded circle denotes the state of the GNSS receiver. The purple and red circles represent the DD pseudorange and carrier-phase factor, respectively. Both the DD pseudorange and carrier-phase measurements connect the state of the GNSS receiver and the position of the reference station. The blue shaded rectangle denotes the Doppler velocity factor which is identical to the one in GNSS positioning (see Section III-A). The green shaded circle denotes the state of the reference station. Be noted that the position of the reference station is fixed throughout the test.

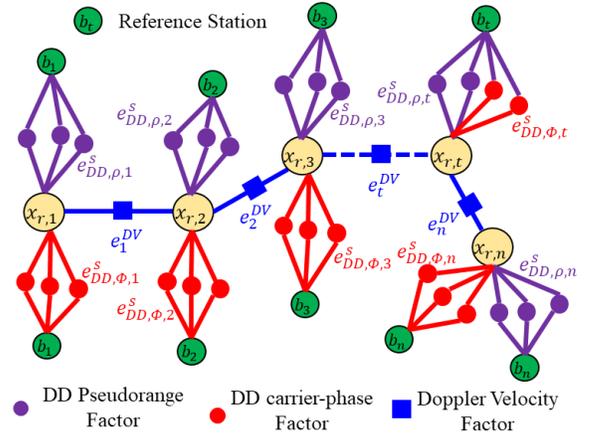

**Fig. 3** The graph structure for estimating the float solution of the GNSS-RTK.

Therefore, the observation model for the DD pseudorange measurement ($\rho_{DD,t}^s$) is expressed as follows:

$$\rho_{DD,t}^s = h_{DD,\rho,t}^s(\mathbf{x}_{r,t}, \mathbf{p}_t^s, \mathbf{p}_t^w, \mathbf{p}_b) + \omega_{DD,\rho,t}^s \qquad (19)$$

$$h_{DD,\rho,t}^s(\mathbf{x}_{r,t}, \mathbf{p}_t^s, \mathbf{p}_t^w, \mathbf{p}_b) = (||\mathbf{x}_{r,t} - \mathbf{p}_t^s|| - ||\mathbf{p}_b - \mathbf{p}_t^s||) - (||\mathbf{x}_{r,t} - \mathbf{p}_t^w|| - ||\mathbf{p}_b - \mathbf{p}_t^w||) \qquad (20)$$

The variable $\omega_{DD,\rho,t}^s$ denotes the noise associated with the $\rho_{DD,t}^s$. The function $h_{DD,\rho,t}^s(*)$ denotes the observation function connecting the state of the GNSS receiver and the DD measurement $\rho_{DD,t}^s$. Therefore, the error factor for the DD pseudorange measurement as follows:

$$||\mathbf{e}_{DD,\rho,t}^s||_{\Sigma_{DD,\rho,t}^s}^2 = ||\rho_{DD,t}^s - h_{DD,\rho,t}^s(\mathbf{x}_{r,t}, \mathbf{p}_t^s, \mathbf{p}_t^w, \mathbf{p}_b)||_{\Sigma_{DD,\rho,t}^s}^2 \quad (21)$$

The variable $\Sigma_{DD,\rho,t}^s$ stands for the covariance associated with the $\rho_{DD,t}^s$. Similarly, the observation model for the DD carrier-phase measurement is expressed as follows:

$$\Phi_{DD,t}^s = h_{DD,\Phi,t}^s(\mathbf{x}_{r,t}, \mathbf{p}_t^s, \mathbf{p}_t^w, \mathbf{p}_b) + \omega_{DD,\Phi,t}^s \quad (22)$$

$$h_{DD,\Phi,t}^s(\mathbf{x}_{r,t}, \mathbf{p}_t^s, \mathbf{p}_t^w, \mathbf{p}_b) = \left(||\mathbf{x}_{r,t} - \mathbf{p}_t^s|| - ||\mathbf{p}_b - \mathbf{p}_t^s||\right) - \left(||\mathbf{x}_{r,t} - \mathbf{p}_t^w|| - ||\mathbf{p}_b - \mathbf{p}_t^w||\right) + \Delta N_{rb,t}^s \quad (23)$$

The variable $\omega_{DD,\Phi,t}^s$ denotes the noise associated with the $\Phi_{DD,t}^s$. The variable $\Delta N_{rb,t}^s$ denotes the DD ambiguity of the carrier-phase measurement. Therefore, the error factor for the DD carrier-phase measurement is as follows:

$$||\mathbf{e}_{DD,\Phi,t}^s||_{\Sigma_{DD,\Phi,t}^s}^2 = ||\Phi_{DD,t}^s - h_{DD,\Phi,t}^s(\mathbf{x}_{r,t}, \mathbf{p}_t^s, \mathbf{p}_t^w, \mathbf{p}_b)||_{\Sigma_{DD,\Phi,t}^s}^2 \quad (24)$$

The variable $\Sigma_{DD,\Phi,t}^s$ stands for the covariance associated with the $\Phi_{DD,t}^s$. Therefore, the objective function for the float solution estimation of GNSS-RTK using FGO is formulated as follows:

$$\chi^* = \arg\min_\chi \sum_{s,t} ||\mathbf{e}_{r,t}^{DV}||_{\Sigma_{r,t}^{DV}}^2 + ||\mathbf{e}_{DD,\rho,t}^s||_{\Sigma_{DD,\rho,t}^s}^2 + ||\mathbf{e}_{DD,\Phi,t}^s||_{\Sigma_{DD,\Phi,t}^s}^2 \quad (25)$$

The variable $\chi^*$ denotes the optimal estimation of the state sets. Therefore, the float solution for GNSS-RTK at the current epoch can be obtained by solving the above objective function (25). After obtaining the float solution of the GNSS-RTK using FGO, an ambiguity resolution algorithm is used to estimate the fixed solution. The variable $\Delta N_{r,t}^s$ should be an integer value when the carrier-phase measurement is free from the noise. This paper makes use of the popular LAMBDA algorithm [32] to solve the integer ambiguity resolution problem. Due to the page's limitations, the detailed presentation of the employed LAMDA algorithm can be found online [2].

## IV. EXPERIMENT RESULTS AND DISCUSSIONS

Two experimental evaluations are conducted to evaluate the performance of the proposed framework. The experimental vehicle is shown on the left side of the following Fig. 4. To validate the effectiveness of the proposed GNSS positioning using FGO, we collected data in a challenging urban canyon 1 (see Fig. 4-(b)) with numerous multipath effects and NLOS receptions. During the test, a u-blox M8T GNSS receiver is used to collect raw single-frequency GPS and BeiDou measurements at a frequency of 1 Hz. Besides, the NovAtel SPAN-CPT, a GNSS RTK/INS (fiber optic gyroscopes) integrated navigation system is used to provide the ground truth. To validate the effectiveness of the proposed GNSS-RTK method using FGO, we collect the other static dataset in a less urbanized area (see Fig. 4-(c), the GNSS-RTK can get fixed in the less urbanized area) using the same sensor setup as the one in urban canyon 1. We run the proposed framework using a high-performance laptop computer with an Intel i7-9750K at 2.60GHz and 32GB RAM. For the GNSS positioning evaluation, we compare the following methods:

(1) **WLS** [4]: GNSS positioning is based on the pseudorange using WLS via RTKLIB.
(2) **EKF**: GNSS positioning based on the integration of pseudorange and velocity from Doppler measurements using the EKF estimator.
(3) **FGO**: GNSS positioning based on the integration of pseudorange and velocity from Doppler measurements using FGO.

For the GNSS-RTK evaluation, we compare the following methods:

(1) **RTK-EKF** [4]: GNSS-RTK positioning based on the DD pseudorange and carrier-phase measurements using the EKF estimator via RTKLIB.
(2) **RTK-FGO**: GNSS-RTK positioning based on the DD pseudorange, Doppler measurement, and carrier-phase measurements using the FGO. Be noted that the integer ambiguity is resolved using the LAMBDA algorithm.

The positioning performance of the listed five methods is evaluated in the east, north, and up (ENU) frame. As the GNSS positioning in the vertical direction is highly unreliable due to the satellite geometry, only the horizontal positioning is evaluated.

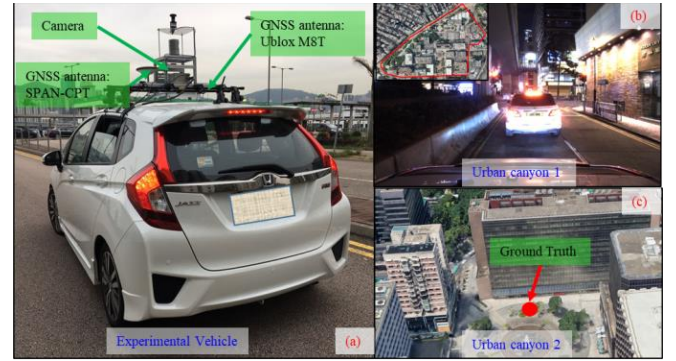

**Fig. 4** The sensor setup of the experimental vehicle and evaluated scenarios. (a) sensor setup of the experimental vehicle. (b) the experimental scene with tall buildings in urban canyon 1. (c) the experimental scene in urban canyon 2.

### A. Evaluation of the Proposed GNSS Positioning

The positioning error of GNSS positioning in the evaluated dataset is shown in the following Table 1. A mean of 17.39 meters is obtained using WLS with a standard deviation (STD) of 16.01 meters. Meanwhile, the maximum error reaches 94.43 meters. After applying the EKF to integrate the pseudorange measurements, and the velocity of the GNSS receiver derived from the Doppler measurement, the mean error decreases to 13.61 meters based on EKF. Meanwhile, the STD is also reduced slightly to 15.19 meters. However, the maximum error still reaches about 89 meters due to the numerous unexpected outlier measurements. The mean error decreases to only 9.45 meters after applying the FGO with a significantly decreased STD of 8.06 meters. Meanwhile, the maximum error decreases to only 31.94 meters. The significantly improved positioning accuracy shows the effectiveness of the proposed framework based on FGO.

---

[2] https://github.com/weisongwen/GraphGNSSLib

Fig. 5 shows the trajectories using three different methods and ground truth. The black curve denotes the ground truth trajectory provided by the SPAN-CPT. The smoother trajectory is achieved using EKF with the help of velocity measurements, compared with the WLS. However, the trajectory can still deviate significantly from the ground truth trajectory in some epochs. With the help of the proposed framework, a smoother trajectory is obtained almost throughout the test.

Table.1 GNSS positioning performance using the three listed methods

| All data | WLS | EKF | FGO |
| --- | --- | --- | --- |
| MEAN (m) | 17.39 | 13.61 | 9.45 |
| STD (m) | 16.01 | 15.19 | 8.06 |
| MAX (m) | 94.43 | 88.97 | 31.94 |
| Availability | 100% | 100% | 100% |

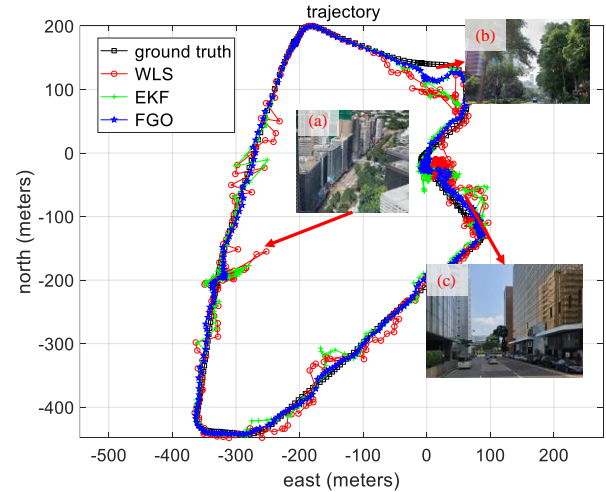

**Fig. 5** Trajectories of three methods WLS (red), EKF (green), and FGO (blue). The x-axis and y-axis denote the east and north directions, respectively.

*B. Evaluation of the Proposed GNSS-RTK*

During the static test in urban canyon 2, the Doppler velocity measurements are employed to connects the consecutive states. The positioning accuracy of GNSS-RTK in the evaluated dataset is shown in the following Table 2. Be noted that the float solution is recorded when the fixed solution is not available. A mean of 2.01 meters is obtained using RTK-EKF with an STD of 0.67 meters. Meanwhile, the maximum error reaches 3.33 meters. The mean error decreases to only 0.64 meters after applying the RTK-FGO with a slightly decreased STD of 0.40 meters. Meanwhile, the maximum error decreases to only 1.70 meters. The improved positioning accuracy shows that the proposed RTK-FGO method can effectively mitigate the effects of the GNSS outlier measurements, leading to improved accuracy.

Fig. 6 shows the trajectories using the two different methods and ground truth. The more accurate trajectory is achieved using RTK-FGO with the help of velocity measurements, compared with the RTK-EKF. We can see from Table 2 that the RTK-EKF gets a fixed rate of 4.4% in the evaluated urban canyon 2. Interestingly, the fixed rate of the proposed RTK-FGO is slightly decreased to 3.8%. The reason is that the proposed RTK-FGO did not consider the cycle slip detection [33] and the ambiguity is solved independently in each epoch.

Table.2 Positioning performance of the GNSS-RTK

| All data | RTK-EKF | RTK-FGO |
| --- | --- | --- |
| MEAN (m) | 2.01 | 0.64 |
| STD (m) | 0.67 | 0.40 |
| MAX (m) | 3.33 | 1.70 |
| Availability | 100% | 100% |
| Fixed Rate | 4.4% | 3.8% |

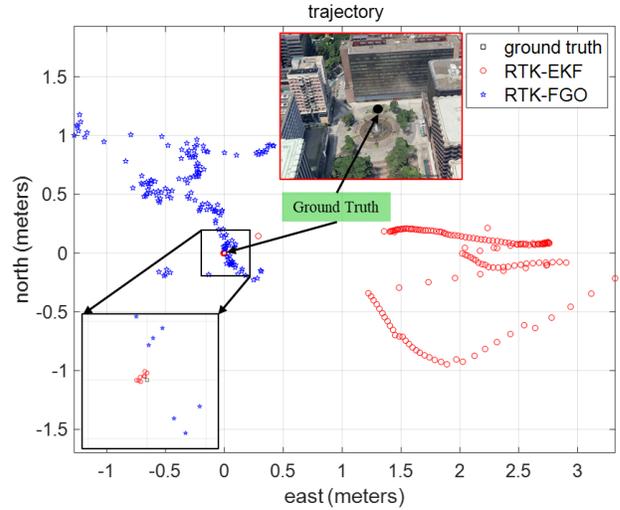

**Fig. 6** Positioning results of two methods RTK-EKF (red dots) and RTK-FGO (blue dots).

V. CONCLUSION AND FUTURE WORK

This paper developed a factor graph-based formulation, that enables the capability of the two most popular positioning methods, the GNSS positioning, and GNSS-RTK. We evaluate the proposed framework using the dataset collected in the urban canyons of Hong Kong. The results show that the proposed method can effectively help to mitigate the effects of GNSS outlier measurements, deriving improved accuracy both in GNSS positioning and GNSS-RTK positioning. The cycle slip detection will be applied to the integer ambiguity resolution to improve the fixed rate in the future. Moreover, achieving a fixed solution for positioning autonomous systems in the urban canyon is still a challenging problem to solve, we will also explore adding more sensors to the proposed framework to increase the fixed rate of GNSS-RTK.

ACKNOWLEDGMENT

The authors acknowledge the ROS, RTKLIB, and the provider of OpenStreetMap.